\documentclass[10pt,twocolumn,letterpaper]{article}

\usepackage{cvpr}              
\definecolor{cvprblue}{rgb}{0.21,0.49,0.74}
\usepackage[pagebackref,breaklinks,colorlinks,allcolors=cvprblue]{hyperref}

\def\paperID{14} 
\def\confName{CVPR}
\def\confYear{2026}

\usepackage{color, colortbl}
\definecolor{LightBlue}{RGB}{212, 250, 252} 
\definecolor{LightGreen}{RGB}{217, 250, 226} 

\title{Design and Behavior of Sparse Mixture-of-Experts Layers in CNN-based Semantic Segmentation}

\author{Svetlana Pavlitska\thanks{Equal contribution} $^{1,2}$, Haixi Fan$^{*1,2}$, Konstantin Ditschuneit$^{2}$, J. Marius Zöllner$^{1,2}$\\
$^{1}$FZI Research Center for Information Technology $^{2}$Karlsruhe Institute of Technology (KIT)\\
{\tt\small pavlitska@fzi.de}
}

\begin{document}
\maketitle
\begin{abstract}
Sparse mixture-of-experts (MoE) layers have been shown to substantially increase model capacity without a proportional increase in computational cost and are widely used in transformer architectures, where they typically replace feed-forward network blocks. In contrast, integrating sparse MoE layers into convolutional neural networks (CNNs) remains inconsistent, with most prior work focusing on fine-grained MoEs operating at the filter or channel levels. In this work, we investigate a coarser, patch-wise formulation of sparse MoE layers for semantic segmentation, where local regions are routed to a small subset of convolutional experts. Through experiments on the Cityscapes and BDD100K datasets using encoder–decoder and backbone-based CNNs, we conduct a design analysis to assess how architectural choices affect routing dynamics and expert specialization. Our results demonstrate consistent, architecture-dependent improvements (up to +3.9 mIoU) with little computational overhead, while revealing strong design sensitivity. Our work provides empirical insights into the design and internal dynamics of sparse MoE layers in CNN-based dense prediction. Our code is available at \url{https://github.com/KASTEL-MobilityLab/moe-layers/}.
\end{abstract}

\section{Introduction}
\label{sec:intro}

Sparse mixture-of-experts (MoE)~\cite{shazeer2017outrageously} architectures have recently demonstrated strong potential within large transformer-based architectures~\cite{fedus2022switch,rajbhandari2022deepspeed,roller2021hash,zoph2022designing,lepikhin2020gshard}. Sparse MoE models fundamentally alter how resources are allocated within a network, concentrating computational power where and when it is most needed. This approach enables significantly larger models without corresponding increases in computational demand during inference, making sparse MoE models a promising solution for efficiently scaling artificial intelligence systems. This adaptability and efficiency make MoE particularly appealing in contexts where computational costs are a critical constraint.

\begin{figure}[t]
\centering
        \includegraphics[width=\columnwidth]{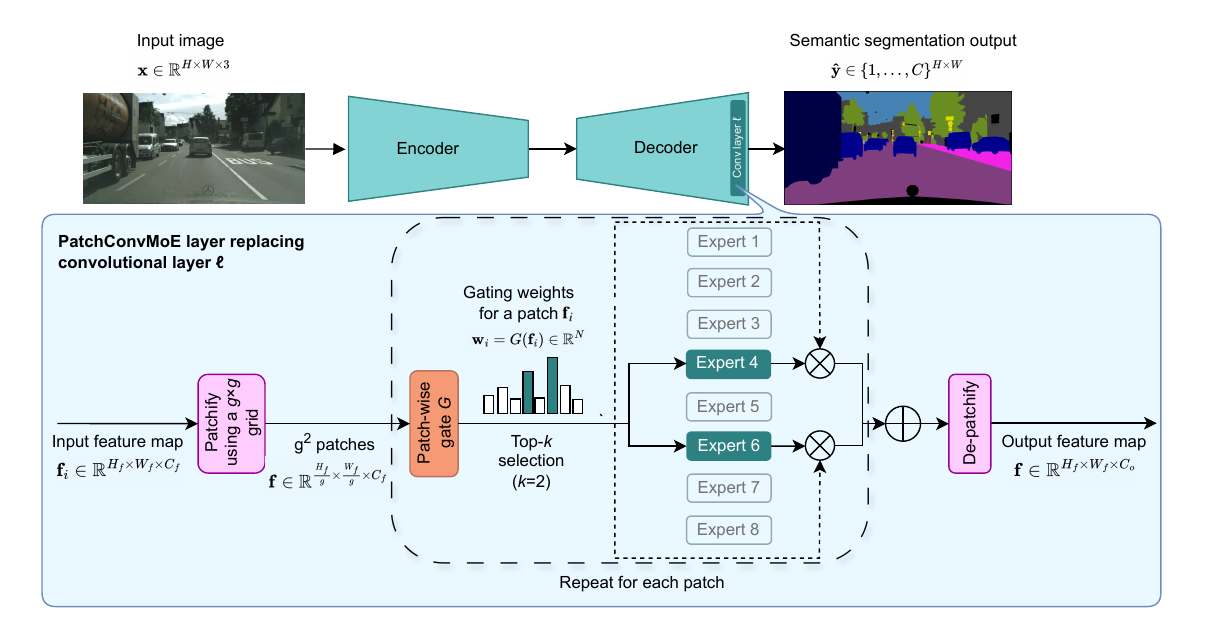}
    \caption{Overview of the studied \textbf{PatchConvMoE} layer within a CNN for semantic segmentation. A standard convolutional layer $\ell$ is replaced with a sparse MoE layer comprising convolutional layers as experts and a patch-wise gating network. The input feature map is split into a $g\times g$ grid of patches. Each patch is routed to a small subset of experts using top-k routing (here: a total of $n=8$ experts with top-2 routing, i.e., $k=2$). Expert outputs are linearly combined and reassembled into the output feature map. }
    \label{fig:concept}
\end{figure}

For convolutional neural networks (CNNs), sparse MoE layers have been mainly applied at the filter~\cite{wang2019deep} or channel~\cite{zhang2023robust,yang2019condconv} level, and less at the layer level~\cite{chen2022towards,pavlitska2023sparsely}, and mainly for the image classification task. Semantic segmentation is inherently more complex, requiring pixel-wise predictions. While sparse MoE layers have been applied to vision transformers for semantic segmentation~\cite{han2024vimoe}, their usage in CNNs for this task has received less attention.  

A \textbf{research gap} thus remains in understanding how to effectively integrate sparse MoE layers into CNNs for dense prediction tasks, such as semantic segmentation, and how these models behave internally. In transformer architectures, MoE layers are a standardized component that replace the feed-forward network blocks and operate naturally at the token or patch level, enabling efficient scaling of model capacity without a proportional increase in computational cost. In contrast, existing MoE formulations for CNNs have been developed primarily for image classification, where image-level routing is sufficient to capture global semantics. However, semantic segmentation requires pixel-level predictions, making such coarse routing inadequate. In this work, we analyze a patch-level routing strategy for sparse MoE layers in CNNs, where the feature map is divided into spatial regions, and each region is routed to a small subset of convolutional experts. Beyond enabling spatial adaptivity at manageable computational cost, this formulation allows us to explicitly characterize and diagnose the behavior of conditional computation in dense prediction networks with a focus on the expert specialization and routing dynamics. 

Our \textbf{contributions} can be summarized as follows: 
\begin{itemize}
    \item We provide a systematic behavioral analysis of sparse MoE layers within CNN-based semantic segmentation models, examining how gating design, routing granularity, expert count, and layer placement affect accuracy, computational overhead, routing stability, and expert specialization.
    \item We evaluate six CNN-based segmentation architectures, including encoder–decoder and backbone-based networks, on the Cityscapes and BDD100K datasets.
\end{itemize}

\section{Related Work}

The MoE was first introduced by Jacobs et al.~\cite{jacobs1991adaptive} in the early 1990s to improve model efficiency and specialization. Their approach relied on multiple expert networks, each specializing in different parts of the input space, and a gating network that learned to assign inputs to the most appropriate expert. This architecture enabled adaptive learning, where experts focused on distinct subtasks, improving interpretability and performance. Works pursuing model-level MoEs mainly focus on the computer vision tasks~\cite{mees2016choosing,valada2016convoluted,ma2018modeling,xu2025limoe,zhang2019learning,pavlitskaya2020using,pavlitskaya2022evaluating}. MoEs at the level of neural network components, which are the focus of this work, have been mostly used in transformers.

\subsection{Sparse MoE Layers for Transformers}

Early MoE models were fully dense, evaluating all experts for every input, resulting in high computational costs. Eigen et al.~\cite{eigen2013learning} suggested integrating MoEs as modular subcomponents within architectures, each with its own learned gating network. A key breakthrough, however, came with the introduction of sparse MoEs by Shazeer et al.~\cite{shazeer2017outrageously}, who proposed a top-k sparsely-gated routing that activates only a small subset of experts per input. This approach preserved model capacity while drastically reducing computational overhead. Lepikhin et al.~\cite{lepikhin2020gshard} extended this concept to transformer models through GShard, demonstrating the feasibility of sparsity in large-scale language models. The idea was further scaled by models such as Switch Transformers~\cite{fedus2022switch} and GLaM~\cite{du2022glam}, which used sparse expert routing to efficiently train trillion-parameter networks. More recent works also include DeepSeek-MoE~\cite{dai2024deepseekmoe}.

In computer vision, sparse MoEs were first applied to vision transformers (ViTs)~\cite{dosovitskiy2020image} by Riquelme et al. in V-MoE~\cite{riquelme2021scaling}, where experts replaced standard feed-forward layers, resulting in significant improvements in scalability with minimal compute overhead. Xue et al.~\cite{xue2022go} extended this approach by incorporating MoE layers across multiple transformer blocks. Hwang et al.~\cite{hwang2023tutel} introduced Tutel, which integrates adaptive sparse experts into hierarchical Swin Transformer~\cite{liu2021swin} backbones to efficiently support large-scale vision tasks. Puigcerver et al.~\cite{puigcerver2023soft} proposed Soft-MoE, replacing top-k gating with a fully differentiable soft routing mechanism that improves convergence and gradient flow. In the remote sensing domain, Rossi et al.~\cite{rossi2025swin2} developed Swin2-MoSE, which combines MoE modules with shifted window attention for high-resolution image restoration. Additionally, Rajbhandari et al.~\cite{rajbhandari2022deepspeed} introduced DeepSpeed-MoE, a training and inference framework that enables the scalable deployment of sparse MoE layers across transformer architectures. Sparse MoEs are thus an established component in transformer architectures.


\subsection{Sparse MoE Layers for CNNs}
MoE layers in CNNs are integrated using diverse strategies, unlike transformers, where they consistently replace feed-forward layers. In CNNs, experts can operate at different granularities: filters, channels, or layers.

Wang et al. were the first to apply sparse MoE layers to CNNs in DeepMoE~\cite{wang2019deep}, where they introduced channel-group experts routed adaptively per input, achieving gains on CIFAR-100 and ImageNet. Similarly, CondConv~\cite{yang2019condconv} incorporated multiple filter sets per convolutional layer, with input-dependent gating leading to improved performance and parameter efficiency. Dynamic Convolution~\cite{chen2020dynamic} used attention over convolution kernels, while SlimConv~\cite{qiu2021slimconv} selected among experts with varying receptive fields to enhance multi-scale learning.

For more modular structures, sparsely-gated MoE modules~\cite{pavlitska2023sparsely} were integrated at the block or layer level, enabling interpretability through semantic expert specialization. AdvMoE~\cite{zhang2023robust} refined granularity further by introducing filter- and channel-level experts within each layer. These were activated via a gating network to form input-dependent pathways, boosting adversarial robustness and efficiency. Chen et al.~\cite{chen2022towards} provided a theoretical analysis of how MoE layers achieve specializations in CNNs for simple vision tasks.

Overall, CNN-based MoEs deliver consistent benefits across vision tasks, with expert placement varying from filter~\cite{yang2019condconv, chen2020dynamic}, channel~\cite{wang2019deep, zhang2023robust}, to block~\cite{pavlitska2023sparsely}. However, to the best of our knowledge, sparse MoEs have not yet been applied to CNNs for semantic segmentation. A related attempt by Han et al.~\cite{han2024vimoe} integrates sparse MoE layers into vision transformers for this task.

\section{Sparse MoE Architecture for CNNs}
We analyze the effect of integrating a sparse MoE layer into CNNs for semantic segmentation.
While the individual components (sparse MoE routing, load balancing, patch-wise feature grouping) have been explored in prior contexts, their combination within CNNs for dense prediction has not been systematically studied. We use a reference configuration, called \textbf{PatchConvMoE}, which serves as the basis for our experiments.

\subsection{MoE Layer Configuration}

To study the effect of sparse expert routing in convolutional networks, we employ convolutional layers as experts, maintaining the spatial inductive bias inherent to CNNs. This choice enables a direct comparison with standard convolutional baselines while allowing for conditional computation through expert selection. Using convolutional experts also ensures compatibility across architectures with minimal modification to existing network designs.

In our configuration, a single 1$\times$1 or 3$\times$3 convolutional layer $\ell$ (depending on the model) is replaced with a \textbf{PatchConvMoE} layer composed of $N$ experts $\{e_1, \dots, e_N\}$ and a gating network $G$. Each expert $e_j$ replicates the structure of the replaced convolutional layer and outputs $C_o$ feature maps. The gate predicts routing weights and activates only the top-$k$ experts for each input, combining their outputs through a weighted sum. This mechanism increases model capacity while keeping the number of active parameters per forward pass close to that of the baseline.

\begin{figure}[h]
    \centering
    \begin{subfigure}{0.82\linewidth}
        \includegraphics[width=\textwidth]{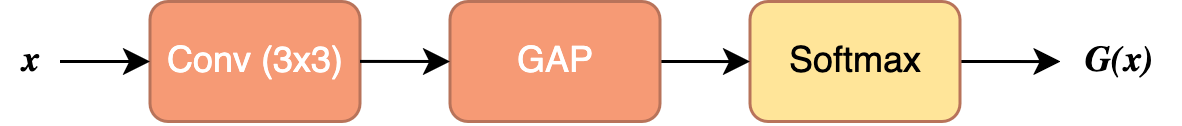}
        \caption{Conv-GAP gate}
    \end{subfigure}
    \begin{subfigure}{\linewidth}
        \includegraphics[width=\textwidth]{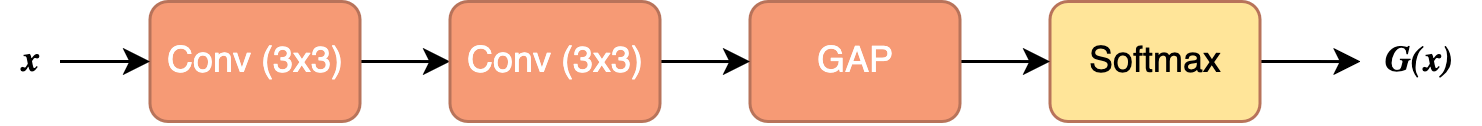}
        \caption{2Conv-GAP gate}
    \end{subfigure}
    \caption{Gate architectures.}
    \label{fig:gates}
\end{figure}

Two gating network variants are analyzed (see Figure~\ref{fig:gates}). The first employs a single 3$\times$3 convolution, followed by global average pooling and a softmax layer, whereas the second uses a deeper two-layer convolutional architecture that can capture more complex spatial dependencies. This comparative setup enables assessment of how gate complexity influences expert specialization and overall segmentation performance.

\subsection{Patch-Level Routing}
Sparse MoE layers in CNNs can operate at different routing granularities. Image-level routing, commonly used in prior work for classification, assigns a single expert configuration to the entire feature map, limiting spatial adaptability. Pixel-level routing, in contrast, allows fine-grained specialization but is computationally prohibitive for high-resolution feature maps. In this work, we therefore focus on an intermediate patch-level routing strategy, which partitions the feature map into spatial regions and routes each patch to a small subset of experts, offering a practical trade-off between spatial flexibility and efficiency.


Given a feature map $\mathbf{f} \in \mathbb{R}^{H_f \times W_f \times C_f}$ as  an input to layer $\ell$, we divide it into a $g \times g$ grid of non-overlapping patches, each of size $\frac{H_f}{g} \times \frac{W_f}{g} \times C_f$, resulting in $p = g^2$ patches $\{\mathbf{f}_1, \dots, \mathbf{f}_p\}$. Each patch $\mathbf{f}_i$ is then processed independently. A gating function $G$ receives the patch as input and produces a sparse weight vector:
\[
G(\mathbf{f}_i) = [g_1(\mathbf{f}_i), \dots, g_N(\mathbf{f}_i)],
\]
which determines the top-$k$ experts selected from a pool of $N$ convolutional experts $\{e_1, \dots, e_N\}$. The output for each patch is computed as a weighted sum of the selected expert outputs:
\[
\mathbf{y}_i = \sum_{j \in \text{TopK}(G(\mathbf{f}_i))} g_j(\mathbf{f}_i) \cdot e_j(\mathbf{f}_i), \quad \mathbf{y}_i \in \mathbb{R}^{\frac{H_f}{g} \times \frac{W_f}{g} \times C_o}.
\]

The patch outputs $\{\mathbf{y}_i\}$ are then reassembled into the output feature map $\mathbf{f}' \in \mathbb{R}^{H_f \times W_f \times C_o}$.

\subsection{Balancing Losses}
A common challenge in training sparse MoE layers is routing collapse~\cite{shazeer2017outrageously}, where the gating network routes most inputs to a small subset of experts, resulting in unbalanced expert usage and reduced generalization. 

To mitigate this effect, several balancing losses have been proposed. The \textbf{importance loss} $\mathcal{L}_{imp}$~\cite{shazeer2017outrageously} penalizes variance in expert importance across a batch to encourage uniform routing. The \textbf{switch loss} $\mathcal{L}_{switch}$~\cite{fedus2022switch} promotes balanced expert activation by directly matching the average routing probability to a uniform prior. The \textbf{entropy loss} $\mathcal{L}_{entropy}$ maximizes the entropy of expert assignments, encouraging diverse expert utilization~\cite{pavlitska2025robust}. 

While some of these losses were originally developed for transformer-based MoEs, their comparative behavior in convolutional architectures has not been systematically analyzed. We therefore include all three variants to examine how balancing losses influence routing stability and expert specialization in CNN-based sparse MoE layers.

\begin{table*}[t]
    \centering
     \caption{Performance of the baselines and models with one convolutional layer replaced with a PatchConvMoE layer in the decoder with 8 experts, top-2 routing, $\mathcal{L}_{entropy}$, and the 3$\times$3 grid for patchification. We highlight the best overall mIoU per architecture in \textbf{bold} and mIoU of models with a PatchConvMoE layer surpassing that of the baseline in \colorbox{LightGreen}{green}. Inference time evaluated on NVIDIA RTX2080Ti.}
    \label{tab:performance}
    \resizebox{\textwidth}{!}{
   
    \begin{tabular}{|r|cc | c | c | cc|}
    \hline  
    \textbf{Architecture} & \multicolumn{2}{c|}{\textbf{Number of parameters}} & \textbf{GFLOPs } & \textbf{Inf. time} &  \multicolumn{2}{c|}{\textbf{mIoU in \%}} \\

     & \textbf{total} & \textbf{active} &  &\textbf{per image, ms} &  \textbf{\texttt{Cityscapes}} & \textbf{\texttt{BDD100K}}\\ \hline
    \rowcolor{LightBlue}
    \multicolumn{7}{|c|}{\texttt{ENet}}\\ 
     Baseline & \multicolumn{2}{c|}{351,660} &8.72 & 12.04 & 51.80  & 52.27\\
     PatchConvMoE, Conv-GAP gate& 352,956 (+0.36\%) & 352,092 (+0.12\%)& 8.73 (+0.11\%)& 25.15 (+108.89\%) & \colorbox{LightGreen}{53.33 (+1.53)} & \colorbox{LightGreen}{53.43 (+1.16)}\\
     PatchConvMoE, 2Conv-GAP gate& 353,548 (+0.54\%) & 352,684 (+0.29\%)& 8.74 (+0.23\%) & 25.20 (+109.30\%) & \colorbox{LightGreen}{\textbf{55.70 (+3.90)}} & \colorbox{LightGreen}{\textbf{53.70 (+1.43)}} \\
     PatchConvMoE, 3Conv-GAP gate& 354,140 (+0.71\%) & 353,276 (+0.46\%)&8.76 (+0.46\%) &  25.26 (+109.80\%) & 50.82 & \colorbox{LightGreen}{53.60 (+1.33)}\\\hline
     
     \rowcolor{LightBlue}
    \multicolumn{7}{|c|}{\texttt{ERFNet}}\\ 
     Baseline & \multicolumn{2}{c|}{2,066,642} &27.70 &  17.96& 60.66 & 54.59 \\
     PatchConvMoE, Conv-GAP gate& 2,157,714  (+4.40\%)& 2,089,410 (+1.10\%) &28.01 (+1.12\%) &  28.64 (+59.47\%)& 60.51 & \colorbox{LightGreen}{54.96 (+0.37)}\\
     PatchConvMoE, 2Conv-GAP gate& 2,158,290 (+4.43\%)& 2,089,986 (+1.13\%) &28.01 (+1.12\%)&  28.69 (+59.74\%)&  \colorbox{LightGreen}{\textbf{61.98 (+1.32)}} & \colorbox{LightGreen}{\textbf{55.32 (+0.73)}} \\
     PatchConvMoE, 3Conv-GAP gate& 2,158,898 (+4.46\%)& 2,090,594 (+1.15\%) & 28.02 (+1.16\%)&  28.75 (+60.08\%)& \colorbox{LightGreen}{60.72 (+0.06)} & \colorbox{LightGreen}{54.84 (+0.25)} \\\hline
     
     \rowcolor{LightBlue}
     \multicolumn{7}{|c|}{\texttt{U-Net}}\\ 
     Baseline & \multicolumn{2}{c|}{4,318,995} &57.90 &   16.83 & 64.94 & 58.50\\
     PatchConvMoE, Conv-GAP gate& 4,338,579 (+0.45\%)& 4,320,147  (+0.02\%)&57.92 (+0.03\%) &22.17 (+31.73\%)& 61.57 & 56.17 \\

     PatchConvMoE, 2Conv-GAP gate& 4,452,627 (+3.09\%)  & 4,342,323 (+0.54\%) &58.21 (+0.54\%)  &  24.21 (+43.85\%)&  63.88 & \colorbox{LightGreen}{\textbf{58.90 (+0.40)}}\\ \hline

      \rowcolor{LightBlue}
    \multicolumn{7}{|c|}{\texttt{LR-ASPP} (\texttt{MobileNetv3})}\\ 
     Baseline & \multicolumn{2}{c|}{3,284,547} &2.09 &  13.93& 66.79 & 61.53\\
     PatchConvMoE, Conv-GAP gate&  3,289,283 (+0.14\%) &  3,286,211 (+0.05\%) & 2.091 (+0.05\%)&  21.22 (+52.33\%)& \colorbox{LightGreen}{67.90 (+1.11)} & \colorbox{LightGreen}{63.84 (+2.31)}\\
     PatchConvMoE, 2Conv-GAP gate& 3,289,875 (+0.16\%) & 3,286,803  (+0.06\%) &2.091 (+0.05\%) &   21.27 (+52.69\%)& \colorbox{LightGreen}{\textbf{68.63} (+1.84)} & \colorbox{LightGreen}{\textbf{64.64 (+3.11)}}\\  \hline 

     \rowcolor{LightBlue}
     \multicolumn{7}{|c|}{\texttt{DeepLabv3+} (\texttt{ResNet50})}\\ 
    Baseline & \multicolumn{2}{c|}{40,351,667} &352.72&  24.19& 77.26 &78.80\\
    PatchConvMoE, Conv-GAP gate& 44,498,867 (+10.28\%) & 40,959,923 (+1.51\%) & 358.03 (+1.51\%)&   33.30 (+37.66\%) & 76.98 & 78.73 \\
    PatchConvMoE, 2Conv-GAP gate& 44,499,459 (+10.28\%) & 40,960,515 (+1.51\%)& 358.04 (+1.51\%)&  33.36 (+37.91\%) & \colorbox{LightGreen}{\textbf{77.61 (+0.35)}} & \colorbox{LightGreen}{\textbf{80.98 (+2.18)}} \\ \hline
    \rowcolor{LightBlue}

    \rowcolor{LightBlue}
    \multicolumn{7}{|c|}{\texttt{PSPNet} (\texttt{ResNet50})}\\ 
     Baseline & \multicolumn{2}{c|}{48,956,262} &356.87 &63.38& 73.05 & 78.12\\
     PatchConvMoE, Conv-GAP gate& 49,061,355 (+0.21\%) & 48,982,535 (+0.05\%) &357.06 (+0.05\%) & 72.49 (+14.37\%)& 72.26 & \colorbox{LightGreen}{79.21 (+1.09)}\\
     PatchConvMoE, 2Conv-GAP gate& 49,061,947 (+0.22\%) & 48,983,127 (+0.05\%) &357.06 (+0.05\%)  &72.54 (+14.45\%)&\colorbox{LightGreen}{\textbf{75.63 (+2.58)}} & \colorbox{LightGreen}{\textbf{81.01 (+2.89)}} \\  \hline 

    \end{tabular}
    }
\end{table*}

\section{Experiments and Evaluation}
We evaluate the architecture described above on two types of CNN models: encoder-decoder models and models with a pre-trained CNN backbone for feature extraction and an attached segmentation head. We demonstrate the impact of various architectural choices on model performance and routing behavior.

\subsection{Experimental Setup}
We evaluate the impact of embedding a PatchConvMoE in \texttt{ENet}~\cite{paszke2016enet}, \texttt{ERFNet}~\cite{romera2018erfnet}, \texttt{U-Net}~\cite{ronneberger2015unet}, \texttt{LR-ASPP}~\cite{howard2019searching}, \texttt{DeepLabv3+}~\cite{chen2018deeplabv3+}, and \texttt{PSPNet}~\cite{zhao2017pyramid} on the \texttt{Cityscapes}~\cite{cordts2016cityscapes} and  \texttt{BDD100k}~\cite{yu2020bdd100k} datasets. For the models with backbones, we use \texttt{ResNet-50}~\cite{he2016resnet} and \texttt{MobileNetv3}~\cite{howard2019searching}. We use $3,475$ fine-grained \texttt{Cityscapes} images with publicly available annotations for $19$ classes. Following the established protocol, we randomly crop patches of size 769$\times$769 px for the \texttt{Cityscapes} and of size 640$\times$640 px for \texttt{BDD100k} dataset. We report mIoU on the official validation subsets.

Our experiments intentionally focus on compact and widely used CNNs to allow controlled comparisons and to analyze the effects of sparse MoE integration under realistic computational constraints. High-end CNNs, such as HRNet~\cite{wang2020deep} variants, achieve higher absolute mIoU on datasets but rely on considerably larger backbones and feature-fusion modules. Since our objective is to understand how sparse expert routing interacts with CNN design rather than to pursue new SotA results, smaller and more interpretable models provide a more suitable experimental setting.

Each model is trained for 200 epochs using SGD with a batch size of 8, a momentum of 0.9, and an initial learning rate of $2\times10^{-2}$ without a learning rate scheduler. For backbone-based models (\texttt{DeepLabv3+}, \texttt{LR-ASPP}, \texttt{PSPNet}), all backbones were pre-trained on \texttt{ImageNet}. Baselines without backbone pre-training demonstrated lower mIoU values. We utilize four types of NVIDIA GPUs (RTX 2080 Ti, RTX 3090, A100, and RTX A6000) to meet the memory requirements of models of varying sizes, with up to 64GB of VRAM.

\subsection{Performance and Efficiency Analysis}

\textbf{Baseline Architecture:}
Embedding sparse MoE layers consistently improves performance across all evaluated architectures and datasets (see Table~\ref{tab:performance}). The largest relative improvement appears for lightweight models such as \texttt{ENet}, reaching \textbf{+3.9~mIoU} on \texttt{Cityscapes} and \texttt{LR-ASPP} with \textbf{+3.11~mIoU} on \texttt{BDD100K}, while larger models such as \texttt{PSPNet} and \texttt{DeepLabv3+} also exhibit consistent, though smaller, absolute gains.

\begin{table*}[t]
    \centering
    \caption{Image-based (i.e., no patchification) vs. patch-based routing and different patch sizes evaluated on \texttt{Cityscapes}. PatchConvMoE layers with 2Conv-GAP gate, other settings follow Table~\ref{tab:performance}.}  
    \label{tab:patch-size}
    \resizebox{1.0\linewidth}{!}{
    
    \begin{tabular}{|r|lr|lr|lr|lr|lr|lr|}
    \hline  
    \textbf{Patch size in px} & \multicolumn{2}{c|}{\textbf{\texttt{ENet}}} & \multicolumn{2}{c|}{\textbf{\texttt{ERFNet}}} & \multicolumn{2}{c|}{\textbf{\texttt{U-Net}}} & \multicolumn{2}{c|}{\textbf{\texttt{LR-ASPP}}} & \multicolumn{2}{c|}{\textbf{\texttt{DeepLabv3+}}} & \multicolumn{2}{c|}
    {\textbf{\texttt{PSPNet}}}  \\ 
      & \textbf{mIoU in \%} & \textbf{Grid size} & \textbf{mIoU in \%} & \textbf{Grid size} & \textbf{mIoU in \%} & \textbf{Grid size} & \textbf{mIoU in \%} & \textbf{Grid size} & \textbf{mIoU in \%} & \textbf{Grid size} & \textbf{mIoU in \%} & \textbf{Grid size}\\ \hline
    No patchification & 31.37 & -& 16.25 & -& 57.95 & -& 55.12 &- &  & - & 64.47 & - \\  \hline
    Patch size 8$\times$8 px& 51.26 & 48$\times$48 & 60.31 & 24$\times$24 & 16.50 & 96$\times$96 &58.93 & 48$\times$48 & 77.14 & 24$\times$24 & 70.44  & 12$\times$12 \\
    
    Patch size 16$\times$16 px & 51.69 & 24$\times$24& 60.24 & 12$\times$12& 18.35 & 48$\times$48 &60.68 & 24$\times$24 & \colorbox{LightGreen}{77.41 (+0.15)} & \colorbox{LightGreen}{12$\times$12} & 72.92 & 6$\times$6 \\
    
    Patch size 32$\times$32 px & \colorbox{LightGreen}{52.97 (+1.17)} & \colorbox{LightGreen}{12$\times$12} & \colorbox{LightGreen}{61.46 (+0.80)} & \colorbox{LightGreen}{6$\times$6} & 20.09 & 24$\times$24 &61.36 & 12$\times$12 & \colorbox{LightGreen}{77.49 (+0.23)} & \colorbox{LightGreen}{6$\times$6} & \colorbox{LightGreen}{\textbf{75.63 (+2.58)}} & \colorbox{LightGreen}{\textbf{3$\times$3}} \\ 
    
    Patch size 64$\times$64 px & \colorbox{LightGreen}{53.56 (+1.76)} & \colorbox{LightGreen}{6$\times$6} & \colorbox{LightGreen}{\textbf{61.98 (+1.32)}} & \colorbox{LightGreen}{\textbf{3$\times$3}} & 28.02 & 12$\times$12 &\colorbox{LightGreen}{66.94 (+0.15)} & \colorbox{LightGreen}{6$\times$6} & \colorbox{LightGreen}{\textbf{77.61 (+0.35)}} & \colorbox{LightGreen}{\textbf{3$\times$3}} & \colorbox{LightGreen}{75.02 (+1.97)} & \colorbox{LightGreen}{1$\times$1}  \\
    
    Patch size 128$\times$128 px & \colorbox{LightGreen}{\textbf{55.70 (+3.90)}} & \colorbox{LightGreen}{\textbf{3$\times$3}} & 46.81 & 1$\times$1 & 56.66 & 6$\times$6 &\colorbox{LightGreen}{\textbf{68.63 (+1.84)}} & \colorbox{LightGreen}{\textbf{3$\times$3}} & \colorbox{LightGreen}{77.56 (+0.30)} & \colorbox{LightGreen}{1$\times$1} & 64.68 & 1$\times$1  \\
    
    Patch size 256$\times$256 px & 50.04 & 1$\times$1 & 24.88 & 1$\times$1 & \colorbox{LightGreen}{\textbf{65.87 (+0.93)}} & \colorbox{LightGreen}{\textbf{3$\times$3}} & 65.47 & 1$\times$1 & 77.24 & 1$\times$1 & 24.88 & 1$\times$1  \\ \hline
    \end{tabular}
    }
\end{table*}

\begin{table*}[t]
    \centering
     \caption{mIoU in \% for different balancing losses and parameter \textit{k} for the  PatchConvMoE layers with 2Conv-GAP gate and top-$k$ routing evaluated on \texttt{Cityscapes}. We highlight the results for the best-performing balancing loss, other settings follow Table~\ref{tab:performance}.}
    \label{tab:losses2}
    \resizebox{\linewidth}{!}{
    \begin{tabular}{|r|cc|cc|cc|cc|cc|cc|}
    \hline  
    \textbf{Loss} & \multicolumn{2}{c|}{\textbf{\texttt{ENet}}} & \multicolumn{2}{c|}{\textbf{\texttt{ERFNet}}} & \multicolumn{2}{c|}{\textbf{\texttt{U-Net}}}& \multicolumn{2}{c|}{\textbf{\texttt{LR-ASPP}}}  & \multicolumn{2}{c|}{\textbf{\texttt{DeepLabv3+}}} & \multicolumn{2}{c|}{\textbf{\texttt{PSPNet}}} \\ 
      & \textbf{top-1}& \textbf{top-2}&\textbf{top-1}& \textbf{top-2}& \textbf{top-1}& \textbf{top-2}& \textbf{top-1}& \textbf{top-2} & \textbf{top-1}& \textbf{top-2} & \textbf{top-1}& \textbf{top-2} \\\hline
     
    No loss & \colorbox{LightGreen}{54.13 (+2.33)} & \colorbox{LightGreen}{55.72 (+3.92)} & 59.58 & 59.44 & 56.59 & 57.02 & \colorbox{LightGreen}{67.47(+0.68)} & \colorbox{LightGreen}{67.11 (+0.32)} & \colorbox{LightGreen}{\textbf{78.05 (+0.79)}} & \colorbox{LightGreen}{\textbf{77.66 (+0.40)}} & \textbf{70.21} & 70.35\\ \hline
    
    $\mathcal{L}_{entropy}$ & \colorbox{LightGreen}{51.92 (+0.12)} & \colorbox{LightGreen}{\textbf{55.70  (+3.90)}}  &\colorbox{LightGreen}{\textbf{62.04 (+1.38)}} & \colorbox{LightGreen}{61.98 (+1.32)}  & 62.83 & 63.88 & \colorbox{LightGreen}{\textbf{68.47 (+1.68)}} & \colorbox{LightGreen}{\textbf{68.63 (+1.84)}} & \colorbox{LightGreen}{77.50 (+0.24)} & \colorbox{LightGreen}{77.61 (+0.35)}  & 62.08 & \colorbox{LightGreen}{75.63 (+2.58)}  \\ 
     $\mathcal{L}_{imp}$ & \colorbox{LightGreen}{\textbf{53.76 (+1.96)}} & \colorbox{LightGreen}{53.29 (+1.49)} & 59.05 & \colorbox{LightGreen}{\textbf{62.12 (+1.46)}}  & \colorbox{LightGreen}{\textbf{66.03 (+1.09)}} & \colorbox{LightGreen}{\textbf{67.21 (+2.27)}} & \colorbox{LightGreen}{67.19 (+0.40)} &\colorbox{LightGreen}{67.35 (+0.56)}& \colorbox{LightGreen}{77.41 (+0.15)} & \colorbox{LightGreen}{77.55 (+0.29)} & 64.97 & \colorbox{LightGreen}{\textbf{76.25 (+3.20)}}  \\
     $\mathcal{L}_{switch}$ & \colorbox{LightGreen}{53.75 (+1.95)} & \colorbox{LightGreen}{54.55 (+2.75)}  & 59.36 & \colorbox{LightGreen}{60.86 (+0.20)}& 64.42 & \colorbox{LightGreen}{65.24 (+0.30)}  & \colorbox{LightGreen}{67.23 (+0.44)}&\colorbox{LightGreen}{67.49 (+0.70)}& \colorbox{LightGreen}{77.30 (+0.04)} &   \colorbox{LightGreen}{77.27 (+0.01)}  & 66.99 & \colorbox{LightGreen}{75.46 (+2.41)} \\ \hline
    \end{tabular}
    }
   
\end{table*}

\begin{table}[t]
      \centering
       \caption{Varying number of experts and parameter \textit{k} for the top-$k$ routing evaluated on \texttt{Cityscapes}. PatchConvMoE layers with 2Conv-GAP gate, other settings follow Table~\ref{tab:performance}.} 
         \label{tab:experts-number}
      \resizebox{\linewidth}{!}{
        \begin{tabular}{|c|c|cccc|}
        \hline  
        \textbf{Number of} & \textbf{Total number of} & \multicolumn{4}{c|}{\textbf{mIoU in \%}}\\ 
        \textbf{experts} & \textbf{parameters} &\textbf{top-1} & \textbf{top-2} & \textbf{top-4} & \textbf{top-8}\\ \hline
        \rowcolor{LightBlue}
        \multicolumn{6}{|c|}{\texttt{ENet}} \\ 
        2 &  352,252 &  \colorbox{LightGreen}{52.09 (+0.29)} & \colorbox{LightGreen}{53.86 (+2.06)}  & - & - \\
        4 &  352,684 & 48.64 & \colorbox{LightGreen}{53.84 (+2.04)}  & \colorbox{LightGreen}{51.97 (+0.17)} & - \\
        8 & 353,548  & 51.68 & \colorbox{LightGreen}{\textbf{55.70 (+3.90)}} & 50.98 & \colorbox{LightGreen}{54.16(+2.36)} \\
        16 & 355,276 & \colorbox{LightGreen}{52.07 (+0.27)} & \colorbox{LightGreen}{52.19 (+0.39)} & \colorbox{LightGreen}{52.05 (+0.25)}  &  50.42 \\\hline
        \rowcolor{LightBlue}
        \multicolumn{6}{|c|}{\texttt{ERFNet}} \\ 
        2 & 2,093,202 & \colorbox{LightGreen}{62.05(+1.39)} & \colorbox{LightGreen}{61.42 (+0.76)} & - & - \\
        4 & 2,157,714  & \colorbox{LightGreen}{62.04(+1.38)} & \colorbox{LightGreen}{61.58 (+0.92)} & \colorbox{LightGreen}{\textbf{62.51 (+1.85)}} & - \\
        8 & 2,236,050 & 60.65 & \colorbox{LightGreen}{61.98(+1.32)} & \colorbox{LightGreen}{60.74 (+0.08)} & \colorbox{LightGreen}{61.43 (+0.77)} \\
        16 & 2,344,338  & 60.33 & 60.42 & 59.90 & \colorbox{LightGreen}{61.49 (+0.83)} \\\hline
        
         \rowcolor{LightBlue}
        \multicolumn{6}{|c|}{\texttt{U-Net}}\\ 
        2 & 4,338,579 & 62.35 & 63.28 & - & - \\
        4 & 4,379,187 & 62.83 & \colorbox{LightGreen}{\textbf{65.87 (+0.93)}} & \colorbox{LightGreen}{65.31(+0.38)} & - \\
        8 & 4,452,627 & 63.30 & 63.88 & 62.32 & 62.57 \\  \hline

        \rowcolor{LightBlue}
        \multicolumn{6}{|c|}{\texttt{LR-ASPP}}\\ 
        2 &  3,286,371 & \colorbox{LightGreen}{\textbf{68.74 (+1.95)}}  & 65.70  & - & - \\
        4 & 3,287,539  & \colorbox{LightGreen}{68.04 (+1.25)}& \colorbox{LightGreen}{67.42 (+0.63)} & \colorbox{LightGreen}{67.04 (+0.25)}  & - \\
        8 & 3,289,875 & \colorbox{LightGreen}{68.47 (+1.68)} & \colorbox{LightGreen}{68.63 (+1.84)}  & \colorbox{LightGreen}{68.32 (+1.53)}  &  65.22 \\  \hline

        \rowcolor{LightBlue}
        \multicolumn{6}{|c|}{\texttt{DeepLabv3+}}\\ 
        2 & 40,960,083 & 76.58 & 76.77 & - & - \\
        4 & 42,139,875 & 76.72 & 76.92 & 74.64 & - \\
        8 & 44,499,459 & \colorbox{LightGreen}{77.50 (+0.24)} & \colorbox{LightGreen}{\textbf{77.61 (+0.35)}} & 75.30 & 75.45 \\ 
        \hline 
        \rowcolor{LightBlue}
        \multicolumn{6}{|c|}{\texttt{PSPNet}}\\ 
        2 & 49,003,033 & 72.89 & \colorbox{LightGreen}{\textbf{76.41 (+3.36})} & - & - \\
        4 & 49,022,671 & 68.62 & \colorbox{LightGreen}{74.83 (+1.78)} & \colorbox{LightGreen}{74.91 (+1.86)} & - \\
        8 & 49,061,947 & 62.08 & \colorbox{LightGreen}{75.63 (+2.58)} & \colorbox{LightGreen}{75.90 (+2.85)} &  \colorbox{LightGreen}{75.45 (+2.40)}\\ 
        \hline 
        \end{tabular}
        }
\end{table}

\begin{table}[t]
    \centering
     \caption{Independent vs. shared experts in the PatchConvMoE layer in \texttt{ENet} with 2Conv-GAP gate and patch size 128$\times$128 px evaluated on \texttt{Cityscapes}, other settings follow Table~\ref{tab:performance}.}
    \label{tab:shared moe}
    \resizebox{1.0\columnwidth}{!}{
    \begin{tabular}{|l r| lc|}
    \hline  
    \textbf{Experts} &  \textbf{Parameters} & \textbf{Routing} & \textbf{mIoU in \%}\\ \hline 

    8 independent experts & 353,548 & top-1 & 51.68 \\
     & & top-2 & \colorbox{LightGreen}{\textbf{55.70  (+3.90)}} \\ \hline
    8 independent + 1 shared experts  & 353,692 & top-1 & 51.45 \\ 
      &  & top-2 & 50.21 \\ \hline
    8 independent + 2 shared experts& 353,836  & top-1 & \colorbox{LightGreen}{54.89  (+3.09)}\\
     &   &  top-2& 50.14\\\hline
    8 independent + 3 shared experts & 353,980 &  top-1 & \colorbox{LightGreen}{52.87 (+1.07)}\\
     &&  top-2 & 50.45\\\hline
    8 independent + 4 shared experts  & 354,124 & top-1& 51.54 \\ 
      &  &  top-2 & \colorbox{LightGreen}{54.49 (+2,69)}\\\hline
    \end{tabular}
   }
\end{table}

\textbf{Number of parameters and inference speed:}
Integrating a PatchConvMoE layer results in a moderate increase in the total number of parameters across all architectures, ranging from around $+0.4\%$ and up to $+4.5\%$ for the encoder–decoder models and up to $+10.3\%$ for the largest backbone-based networks (see Table~\ref{tab:performance}). Because only a small subset of experts is active for each input, the number of parameters effectively used during inference remains close to that of the baseline, reaching up to \textbf{$+1.51\%$}. The increase in GFLOPs is also small, up to $+1.51\%$. Training time increased up to 30\% compared to the baseline, depending on the architecture. The increase in inference time is larger for lightweight models, reaching $+14.45\%$ for \texttt{PSPNet}. Note that the higher accuracy does not stem solely from increased capacity, because models with top-1 routing using a single convolutional layer during inference, as in the baseline, still achieve higher mIoU. These results show that sparse expert routing scales efficiently, providing consistent improvements in segmentation on both datasets without compromising computational performance. 

\subsection{Ablations for MoE Layer Components}
In the following, we discuss the impact of architectural choices for PatchConvMoE layer components on the model performance. Due to space constraints, we perform detailed ablations on \texttt{Cityscapes}, which reflect consistent trends observed for \texttt{BDD100K}.

\begin{table*}[t]
    \centering 
    \caption{Number and positions of MoE layers replacing convolutional layers evaluated on \texttt{Cityscapes}. We use architecture-specific layer names to facilitate reproducibility. PatchConvMoE layers with 2Conv-GAP gate, other settings follow Table~\ref{tab:performance}.}
    \label{tab:layer-pos}
            \resizebox{\linewidth}{!}{
    \begin{tabular}{|c|lr  rc|}
    \hline  
    \textbf{Num MoE} & \textbf{MoE Layer position} & \textbf{Patch} & \textbf{Total number of} & \textbf{mIoU}\\ 
    \textbf{layers} &  & \textbf{size, px} & \textbf{parameters} & \textbf{in \%}\\ \hline 
    \rowcolor{LightBlue}
    \multicolumn{5}{|c|}{\texttt{ENet}}\\
    1& 3$\times$3 conv layer in first encoder regular block (\textit{regular1\_1}) & 64$\times$64 & 369,532 & \colorbox{LightGreen}{53.56 (+1.76)} \\ 
    1& 3$\times$3 conv layer in bridge regular block (\textit{regular3\_4}) &32$\times$32 & 419,068 & \colorbox{LightGreen}{54.80 (+3.00)}\\ 
    1& Last 3$\times$3 conv layer in the decoder (\textit{regular5\_1}) &128$\times$128 & 353,548 & \colorbox{LightGreen}{\textbf{55.70 (+3.90)}}\\ 
    2& \textit{regular3\_4} and \textit{regular5\_1} &32$\times$32,128$\times$128 & 420,956& 36.46\\
    3&  Three conv layers in \textit{regular5\_1} &128$\times$128& 389,292 & 32.31 \\
    3 & Conv layers in \textit{regular1\_1} and \textit{regular3\_4} and \textit{regular5\_1} &64$\times$64, 32$\times$32,128$\times$128 & 438,828 & 34.81\\
    \hline
    \rowcolor{LightBlue}
     \multicolumn{5}{|c|}{\texttt{ERFNet}}\\
    1& First 3$\times$3 conv layer in encoder after downsampling (\textit{encoder.layer1}) &64$\times$64& 2,108,610 & \colorbox{LightGreen}{61.05 (+0.39)} \\
    1&  3$\times$3 conv layer in the middle of the bridge (\textit{encoder.layer10}) &32$\times$32& 2,224,002  & 59.95 \\
    1&  Last 3$\times$3 conv layer in decoder (\textit{decoder.layer2}) &64$\times$64& 2,157,714 & \colorbox{LightGreen}{\textbf{61.98 (+1.32)}}\\
    2&  Last two 3$\times$3 conv layer in decoder (\textit{decoder.layer2}) &64$\times$64 & 2,249,970& 60.61\\
    3&  Conv layers in \textit{encoder.layer1} (\textit{encoder.layer10}) \textit{decoder.layer2}& 64$\times$64, 32$\times$32,,64$\times$64& 2,604,738 & 59.96\\\hline
   \rowcolor{LightBlue}
    \multicolumn{5}{|c|}{\texttt{U-Net}}\\
    1& First 3$\times$3 conv layer in encoder after downsampling, \textit{down1.maxpoolconv} &128$\times$128& 4,434,499 & 62.83 \\
    1&  Last 3$\times$3 conv layer in encoder (\textit{down4.maxpoolconv}) &16$\times$16 &  6,097,683 & 22.01\\
    1&  Last 3$\times$3 conv layer in decoder (\textit{up4.conv.doubleconv}) &256$\times$256 & 4,379,187 & \colorbox{LightGreen}{\textbf{65.87 (+0.93)}} \\
    2& Last two 3$\times$3 conv layer in decoder (\textit{up4.conv.doubleconv}) &256$\times$256 &  4,452,915&  46.89 \\
    3& Conv layers in \textit{down1.maxpoolconv} and \textit{down4.maxpoolconv} and \textit{up4.conv.doubleconv} &128$\times$128,16$\times$16, 256$\times$256 & 6,885,939 & 28.02 \\ \hline
    
      \rowcolor{LightBlue}
    \multicolumn{5}{|c|}{\texttt{LR-ASPP}}\\
    1& 3$\times$3 conv layer at the beginning of backbone \textit{(trunk.block1.1)} &128$\times$128& 3,293,779 &  63.68\\
    1& 3$\times$3 conv layer at the end of backbone \textit{(trunk.block4.1)} &32$\times$32 & 3,375,859  &45.66 \\
    1&  First 1$\times$1 conv layer in decoder (\textit{convs2}) &128$\times$128 & 3,289,875 & \colorbox{LightGreen}{\textbf{68.63 (+1.84)}} \\
    2& Last two 1$\times$1 conv layer in decoder (\textit{convs2, convup3}) &128$\times$128 &  3,445,347 &  64.89 \\
    3& Conv layers at the beginning and end of backbone \textit{(trunk.block1.1, trunk.block4.1)} and  in decoder (\textit{convs2}) &128$\times$128,32$\times$32, 128$\times$128 & 3,391,499 & 34.57 \\ \hline
    
    \rowcolor{LightBlue}
    \multicolumn{5}{|c|}{\texttt{DeepLabv3+}}\\
    1&  3$\times$3 conv layer at the beginning of backbone \textit{(backbone.layer1.2)}& 64$\times$64 & 40,614,915 & 73.41\\
    1&  3$\times$3 conv layer at the end of backbone \textit{(backbone.layer4.2)}& 16$\times$16& 56,904,195 & 48.41 \\
    1&  Last 3$\times$3 conv layer in decoder \textit{(decoder.lastconv.3)} &64$\times$64& 44,499,459 & \colorbox{LightGreen}{\textbf{77.61 (+0.35)}} \\
    2&  Last two 3$\times$3 conv layers in decoder \textit{(decoder.lastconv.0/3)}& 64$\times$64 & 45,277,059 & 73.87\\
    3&  Last two 3$\times$3 conv layers in decoder, last 1$\times$1 conv layer in decoder & 64$\times$64,64$\times$64,64$\times$64  & 45,330,264 & 45.62\\\hline

    \rowcolor{LightBlue}
    \multicolumn{5}{|c|}{\texttt{PSPNet}}\\
    1& 3$\times$3 conv layer at the beginning of backbone \textit{(layer1.2.conv2)} &64$\times$64& 49,219,510 & \colorbox{LightGreen}{75.56 (+2.51)} \\
    1&  3$\times$3 conv layer at the end of backbone \textit{(layer4.2)} &16$\times$16 &  65,567,832 & 66.01\\
    1&  Last 1$\times$1 conv layer before upsampling in decoder (\textit{cls.4}) &32$\times$32 & 49,061,947 & \colorbox{LightGreen}{\textbf{75.63 (+2.58)}} \\
    1&  Last 3$\times$3 conv layer before upsampling in decoder (\textit{aux.0}) &32$\times$32 & 65,545,654 & \colorbox{LightGreen}{75.02 (+1.98)} \\
    2& Last two 1$\times$1 conv layer in decoder (\textit{cls.4, aux.4}) &32$\times$32 &  49,272,715& \colorbox{LightGreen}{75.03 (+1.99)} \\
    3& 3$\times$3 conv layer at the beginning and end of backbone \textit{(layer1.2, layer4.2)} and in decoder (\textit{aux.4}) &64$\times$64,16$\times$16, 32$\times$32 & 65,862,107 & 59.02 \\ \hline
    
    \end{tabular}
    }
\end{table*}

\textbf{Gate Architecture:} The 2Conv-GAP gate with two instead of one convolutional layer leads to an increase in mIoU performance compared to the baseline for all models (see Table~\ref{tab:performance}). Additionally, 2Conv-GAP generally yields a higher mIoU than Conv-GAP. We exemplarily evaluate further increasing the number of convolutional layers in the gate to three (3Conv-GAP) for \texttt{ENet} and \texttt{ERFNet}, but it leads to worse performance compared to 2Conv-GAP. An overly complex gate might therefore obstruct learning. The increase in the total number of parameters in the model is insignificant between the gate architectures.


\textbf{Patchification:} Patch-wise routing leads to higher mIoU compared to image-wise routing. The image-wise routing consistently performs worse than the baseline (see Table~\ref{tab:patch-size}). The optimal grid size and correspondingly patch size for each architecture depend on the feature map size before the PatchConvMoE layer. While the optimal patch size differs across the architectures, it mostly corresponded to the 3$\times$3 grid size for all models. Therefore, we further use models with the optimal patch size corresponding to a 3$\times$3 grid for all experiments.

\textbf{Balancing loss:} The entropy loss and the importance loss lead to the best results(see Table~\ref{tab:losses2}). Because the switch loss is typically used in conjunction with top-1 routing, we also evaluate $k=1$, but it does not demonstrate performance improvement. Completely omitting the balancing loss has led to improved MoE performance for some models. Without balancing losses, however, expert usage collapsed to a single dominant expert, preventing expert specialization.

\textbf{Number of experts:} The best-performing architecture is typically the one with four or eight experts. Further increasing the number of experts does not result in a corresponding increase in performance (see Table~\ref{tab:experts-number}).  Unlike MoE applications in transformers, where models are often expanded by more than 10$\times$~\cite{fedus2022switch}, we achieve only approximately 2$\times$ scaling. Depending on the architecture, increasing the number of experts beyond 8 or 16 can lead to a significant increase in the number of parameters, making the architecture less appealing due to higher computational costs.

\textbf{Number of activated experts ($k$):} 
Our experiments (see Table~\ref{tab:experts-number}) show that top-2 routing usually provides better accuracy than top-1. However, continuing to activate more experts in training does not sustain this trend.  Since we position the PatchConvMoE layer at the end of the model, activating more experts may exacerbate training instability and increase the difficulty of loss convergence. Furthermore, these experiments stress the importance of sparsity. For $n$ experts, models without sparsity, i.e., with $k=n$, in most cases achieve worse results than sparse experts, where $k<n$.

\textbf{Independent vs. shared experts:} Motivated by the recent success of shared experts for sparse MoE layers in vision transformers~\cite{dai2024deepseekmoe,xue2022go,han2024vimoe}, we evaluate the impact of adding a shared expert exemplarily for the \texttt{ENet} (see Table~\ref{tab:shared moe}). None of the variants with shared experts reach the performance of a model with independent experts, although several models can beat the baseline in terms of mIoU. Since no previous works have explored the effect of shared experts in CNNs, these results suggest that shared experts may be less beneficial for CNNs than for transformers.

\subsection{Ablations for the Number and Position of MoE Layers}
We evaluate the placement of MoE layers at three different positions: the encoder's last convolutional layer as well as the decoder's first and last convolutional layers (see Table~\ref{tab:layer-pos}). Note that the optimal patch size varies for MoE layers placed in different positions within the architecture.

The placement of the MoE layer in the decoder is evidently more beneficial than in the encoder, likely because decoder layers are more directly involved in refining the segmentation output based on the learned features. The positions with a relatively higher number of parameters (the last 3$\times$3 convolutional layer in the decoder) correlate with slightly better performance in terms of mIoU, potentially indicating that more parameters in the MoE layers contribute positively to the segmentation task to a certain extent. 

Replacing multiple convolutional layers with MoE layers leads to significantly worse segmentation performance. Sparse routing introduces conditional computation, leading to high variance across training runs. With multiple MoE layers, the network becomes unstable. A single optimally placed MoE layer thus offers the benefits of adaptive computation without the instability associated with excessive sparsity. This is different from transformers, where typically all FFN layers are replaced with MoE layers.

\subsection{Routing and Interpretability Analysis}
In the following, we analyze expert use with quantitative and qualitative methods and assess the extent to which routing collapse is pronounced.

\textbf{Quantitative routing analysis} (see Table~\ref{tab:routing-analysis}) uses the normalized routing entropy (NRE) and top expert concentration (TEC). NRE is computed by deriving the distribution of how often each expert is selected. NRE thus measures how evenly routing decisions are spread across experts, with lower values indicating routing collapse. TEC is computed by measuring the proportion of patches assigned to the most frequently selected experts and reflects the degree of routing dominance, with higher values indicating stronger routing collapse. The analysis shows that routing diversity strongly depends on both the balancing loss and the underlying architecture. The switch loss consistently yields the highest NRE and comparatively low TEC across all models, indicating stable and well-distributed expert utilization. In contrast, the importance loss leads to substantially lower NRE and markedly higher TCE values, particularly for lightweight architectures, revealing pronounced expert dominance. The entropy loss generally maintains high NRE but still exhibits elevated TCE in several models, suggesting partial imbalance despite globally diverse routing. Larger backbone-based models show more stable routing across losses, while smaller models are more sensitive to the choice of loss.

\begin{table}[t]
    \centering
    \caption{Routing collapse during inference for the PatchConvMoE models with 2Conv-GAP gate and 8 experts on the \texttt{Cityscapes} data: normalized routing entropy (NRE), where lower values mean routing collapse, and top expert concentration (TEC), where higher values mean routing collapse. Cases with the most stable routing are marked \textbf{bold}.}
    \label{tab:routing-analysis}
    \resizebox{\columnwidth}{!}{
    \begin{tabular}{|r|cc|cc|cc|}
    \hline  
    \textbf{Model} & \multicolumn{2}{c|}{$\mathcal{L}_{switch}$}  & \multicolumn{2}{c}{$\mathcal{L}_{imp}$} & \multicolumn{2}{c|}{$\mathcal{L}_{entropy}$}\\  
    & \textbf{NRE} & \textbf{TEC}& \textbf{NRE} & \textbf{TEC}& \textbf{NRE} & \textbf{TEC}\\\hline
     \texttt{ENet}& 0.9255 &0.4394&0.5764&0.7425&0.8877 &0.5024\\\hline
     
     \texttt{ERFNet}&0.9686 & 0.5183&0.8057 &0.7790& 0.8408&0.5563\\\hline
     
     \texttt{U-Net}&0.9887 &0.4409&\textbf{0.9880} &0.4894&\textbf{0.9874}&0.5330\\ \hline

     \texttt{LR-ASPP}& \textbf{0.9951}&\textbf{0.2951} & 0.7701&0.6193& 0.9468 &0.4931\\  \hline 

    \texttt{DeepLabv3+} &0.9660 &0.4379&0.9606 &0.3888&0.8629& \textbf{0.4724}\\ \hline

     \texttt{PSPNet} & 0.9738 &0.3537& 0.9686&\textbf{0.3595} &0.9177 &0.5189\\  \hline

    \end{tabular}
    }
\end{table}

\begin{figure}[h]
    \centering
    \resizebox{\columnwidth}{!}{
    \begin{tabular}{ccc}
        \includegraphics[width=0.23\textwidth]{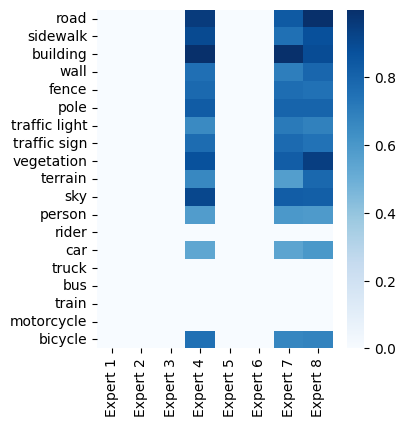}&
        \includegraphics[width=0.23\textwidth]{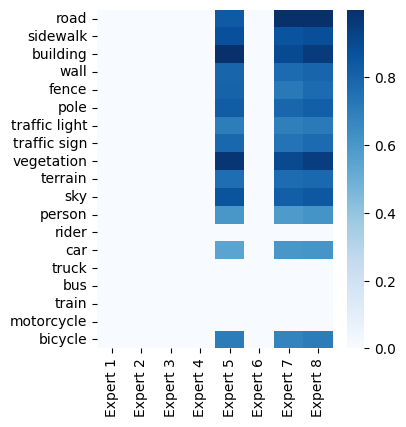}&
        \includegraphics[width=0.23\textwidth]{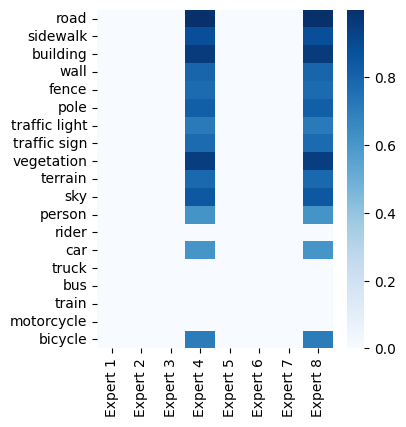}\\ 
        a) \texttt{ENet}, $\mathcal{L}_{switch}$  & b) \texttt{ENet}, $\mathcal{L}_{imp}$ & c) \texttt{ENet}, $\mathcal{L}_{entropy}$ \\ 
        
        \includegraphics[width=0.23\textwidth]{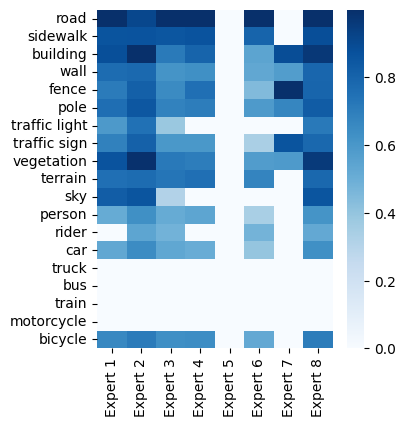}&
        \includegraphics[width=0.23\textwidth]{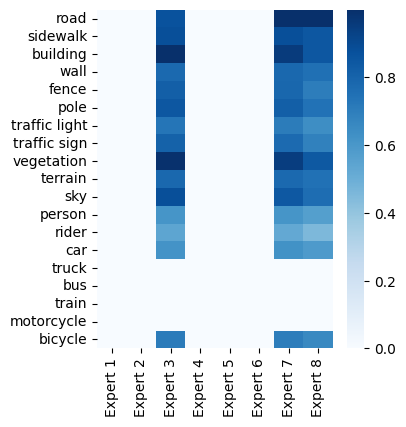}&
        \includegraphics[width=0.23\textwidth]{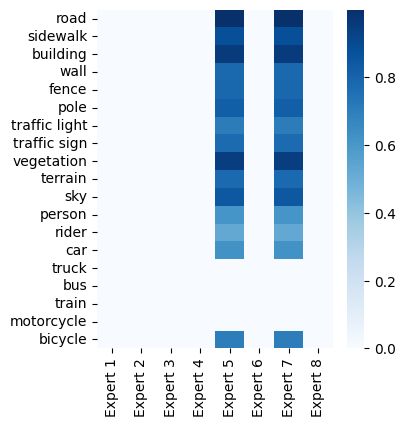}\\
        d) \texttt{ERFNet}, $\mathcal{L}_{switch}$ & e) \texttt{ERFNet}, $\mathcal{L}_{imp}$ & f) \texttt{ERFNet}, $\mathcal{L}_{entropy}$ \\ 
        
        \includegraphics[width=0.23\textwidth]{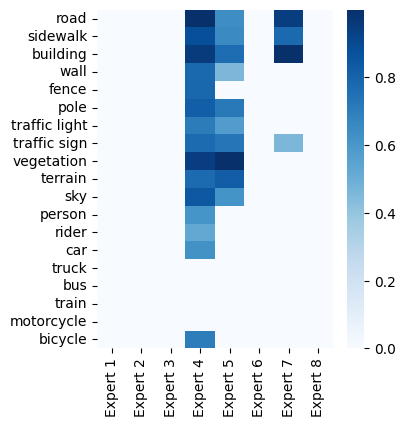}&
        \includegraphics[width=0.23\textwidth]{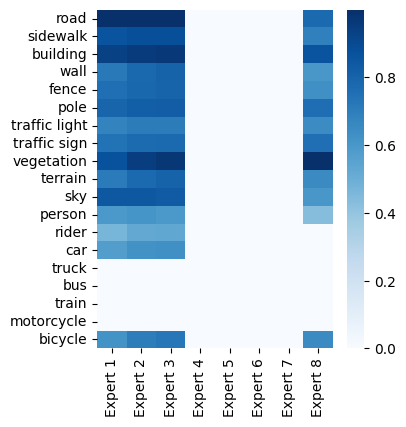}&
        \includegraphics[width=0.23\textwidth]{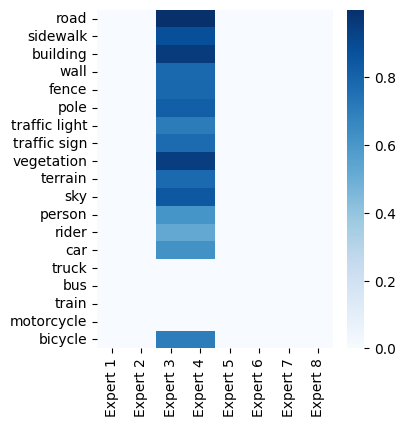}\\
        g) \texttt{U-Net}, $\mathcal{L}_{switch}$  & h) \texttt{U-Net}, $\mathcal{L}_{imp}$ & i) \texttt{U-Net}, $\mathcal{L}_{entropy}$ \\ 
        
        \includegraphics[width=0.23\textwidth]{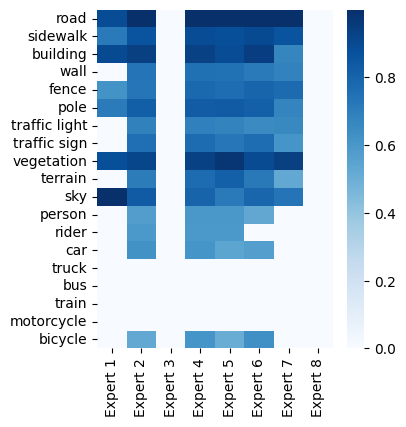}&
        \includegraphics[width=0.23\textwidth]{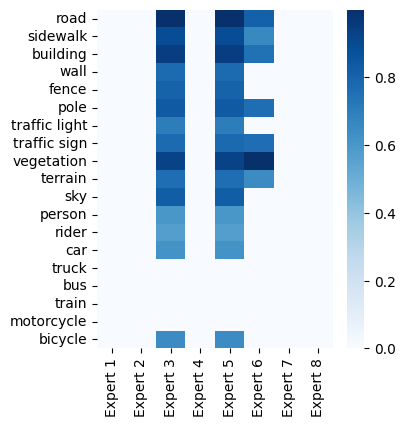}&
        \includegraphics[width=0.23\textwidth]{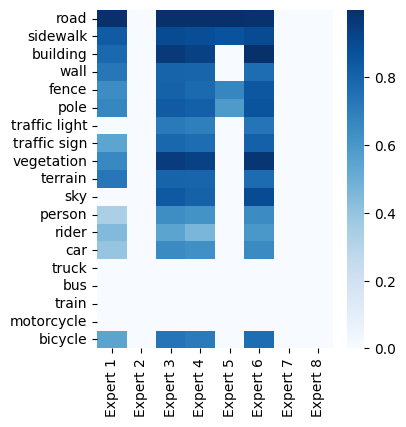}\\
       j) \texttt{LR-ASPP}, $\mathcal{L}_{switch}$  & k) \texttt{LR-ASPP}, $\mathcal{L}_{imp}$ & l) \texttt{LR-ASPP}, $\mathcal{L}_{entropy}$ \\ 
        
        \includegraphics[width=0.23\textwidth]{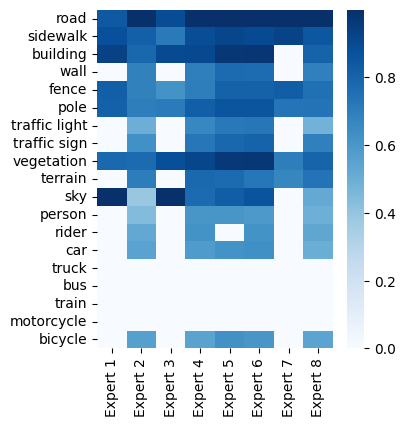}&
        \includegraphics[width=0.23\textwidth]{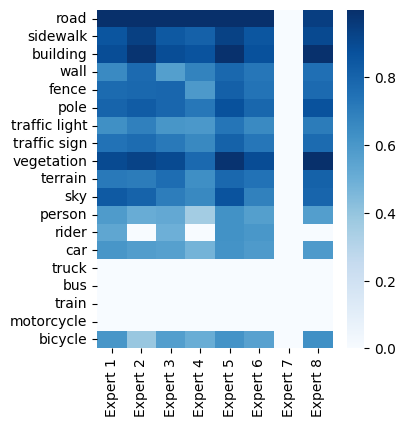}&
        \includegraphics[width=0.23\textwidth]{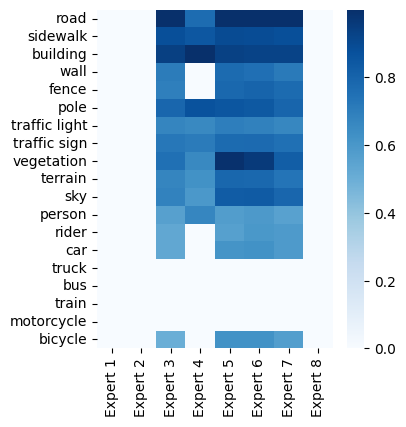}\\
        m) \texttt{DeepLabv3+}, $\mathcal{L}_{switch}$ & n) \texttt{DeepLabv3+}, $\mathcal{L}_{imp}$ & o) \texttt{DeepLabv3+}, $\mathcal{L}_{entropy}$ \\ 
    \end{tabular}
    }
    \caption{Routing heatmaps during inference for models with one convolutional layer replaced with a PatchConvMoE layer in the decoder with 8 experts, 2Conv-GAP gate, top-2 routing, $\mathcal{L}_{entropy}$, and the 3$\times$3 grid for patchification evaluated on \texttt{Cityscapes}.}
    \label{fig:routing-heatmaps}
\end{figure}

\textbf{Expert–class heatmaps} (see Figure~\ref{fig:routing-heatmaps}) show how often each class–expert pair is used during test-time routing. 
Expert utilization varies notably with the balancing loss: switch loss yields the most diverse routing with several active experts, importance loss causes partial collapse with about three dominant experts, and entropy loss shows the strongest collapse, with only a few experts active across most classes. 
Large static classes like road or building dominate routing under collapse, whereas smaller objects such as person, rider, and traffic sign exhibit clearer differentiation under switch loss. 
Across models, \texttt{ENet} is largely insensitive to the loss type, while deeper networks (\texttt{ERFNet}, \texttt{U-Net}, \texttt{DeepLabv3+}) achieve richer, more class-selective expert usage, indicating that broader participation scales with model capacity.

\begin{figure}[h]
    \centering
    \resizebox{\columnwidth}{!}{
    \begin{tabular}{ccc}
        \includegraphics[width=0.23\textwidth]{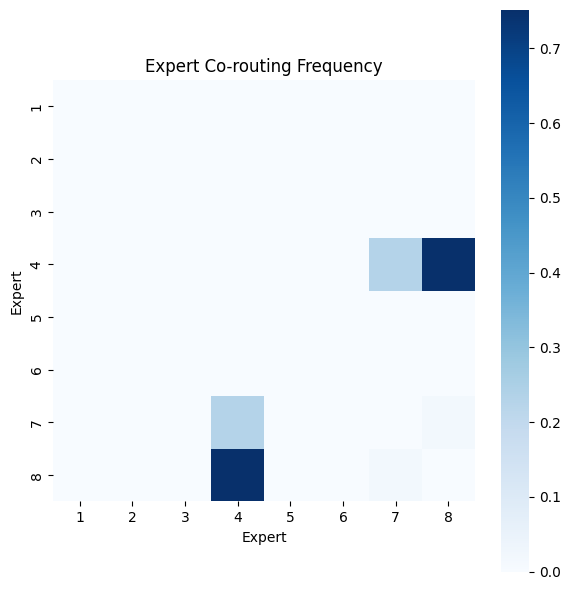}&
        \includegraphics[width=0.23\textwidth]{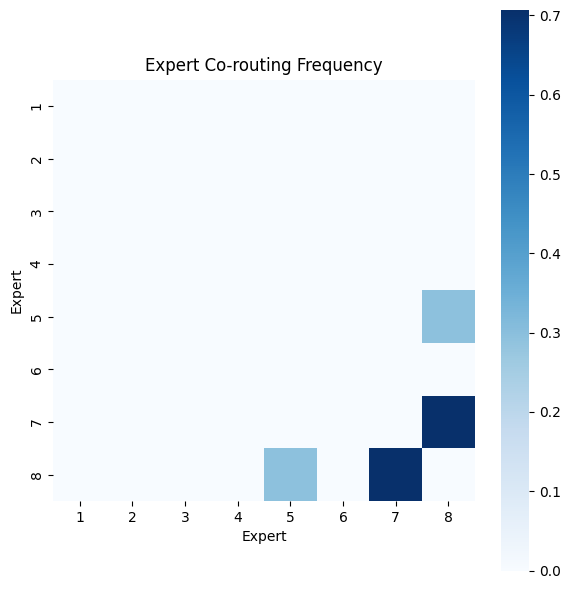}&
        \includegraphics[width=0.23\textwidth]{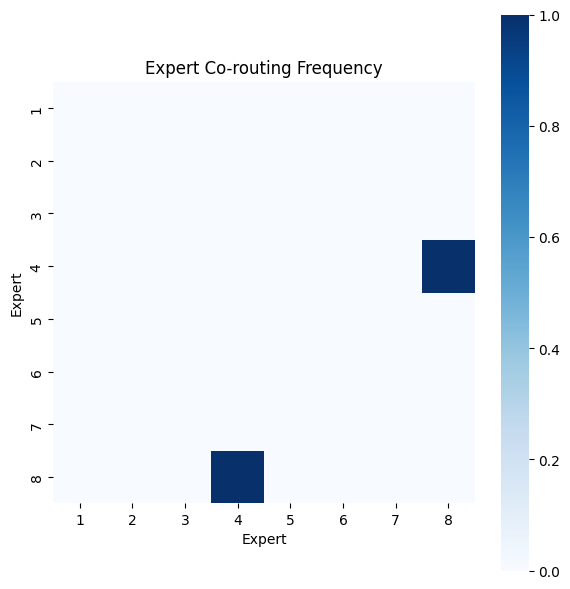}\\
        a) \texttt{ENet}, $\mathcal{L}_{switch}$  & b) \texttt{ENet}, $\mathcal{L}_{imp}$ & c) \texttt{ENet}, $\mathcal{L}_{entropy}$ \\ 
        
        \includegraphics[width=0.23\textwidth]{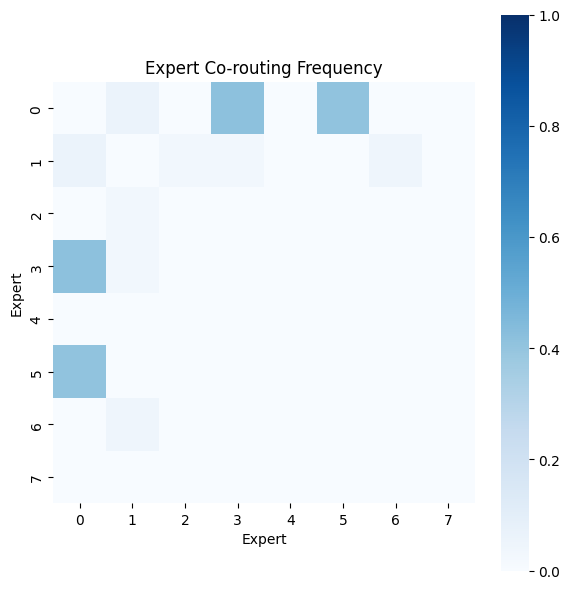}&
        \includegraphics[width=0.23\textwidth]{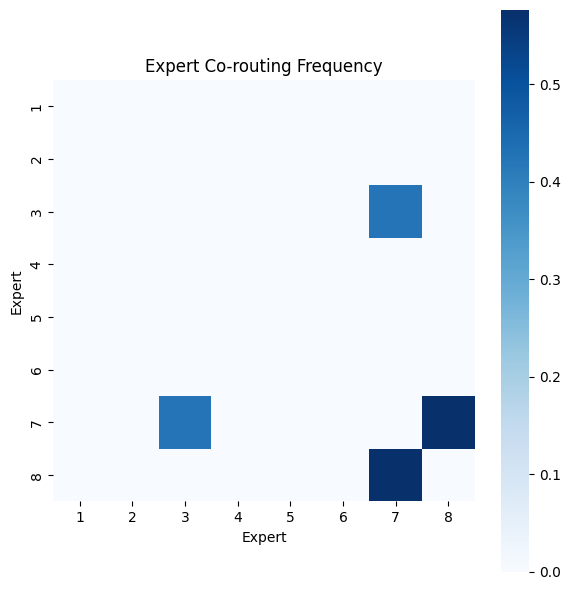}&
        \includegraphics[width=0.23\textwidth]{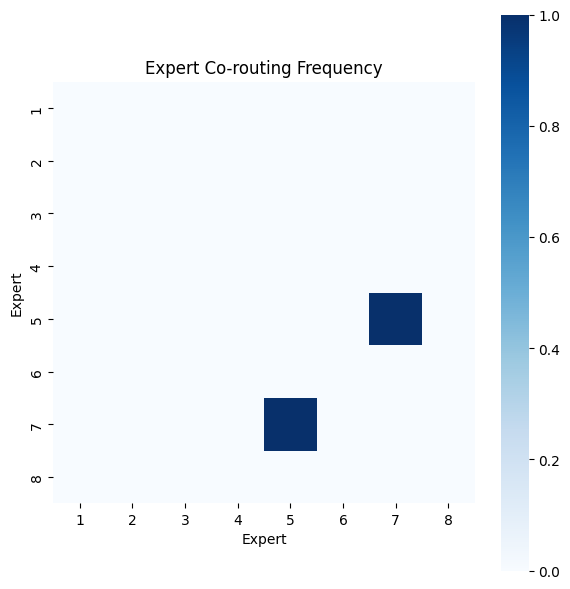}\\
        d) \texttt{ERFNet}, $\mathcal{L}_{switch}$  & e) \texttt{ERFNet}, $\mathcal{L}_{imp}$ & f) \texttt{ERFNet}, $\mathcal{L}_{entropy}$ \\ 
        
        \includegraphics[width=0.23\textwidth]{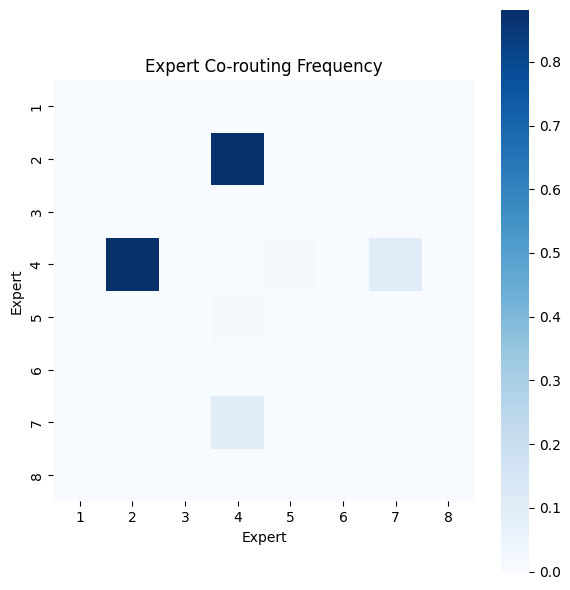}&
        \includegraphics[width=0.23\textwidth]{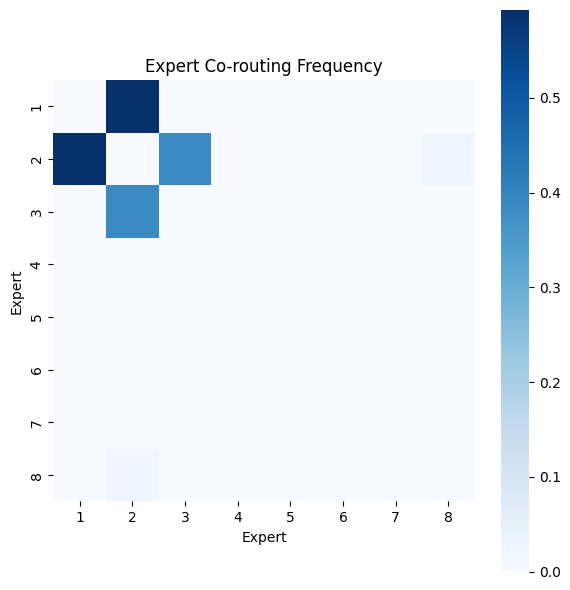}&
        \includegraphics[width=0.23\textwidth]{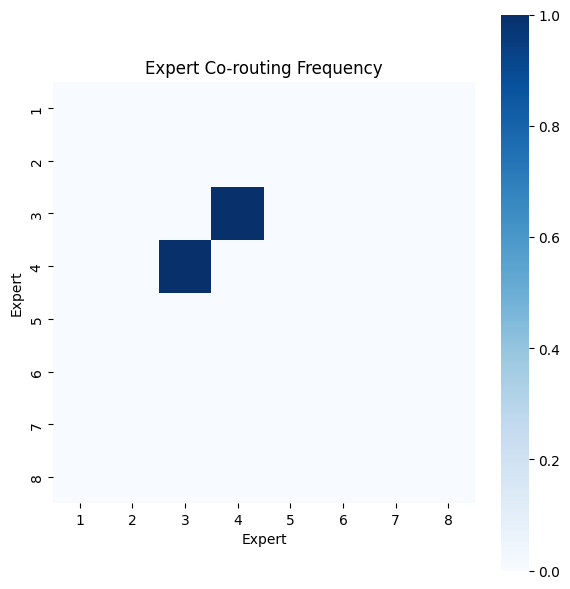}\\
        g) \texttt{U-Net}, $\mathcal{L}_{switch}$  & h) \texttt{U-Net}, $\mathcal{L}_{imp}$ & i) \texttt{U-Net}, $\mathcal{L}_{entropy}$ \\ 
        
        \includegraphics[width=0.23\textwidth]{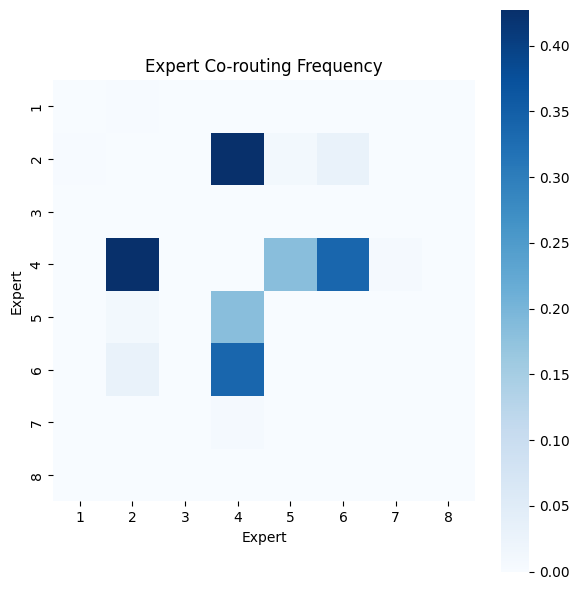}&
        \includegraphics[width=0.23\textwidth]{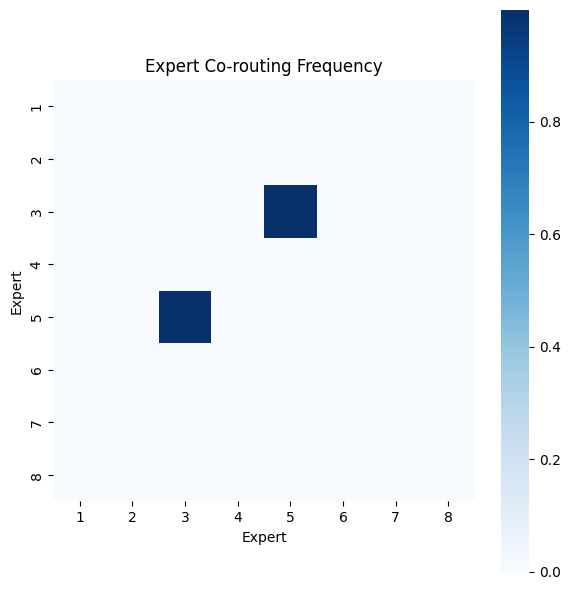}&
        \includegraphics[width=0.23\textwidth]{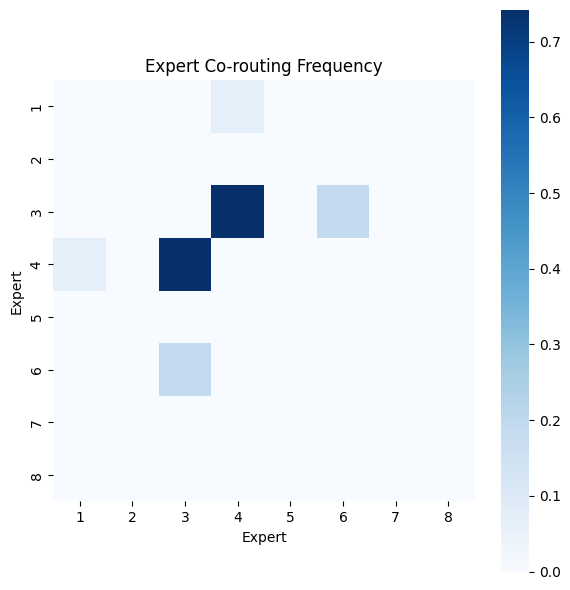}\\
        j) \texttt{LR-ASPP}, $\mathcal{L}_{switch}$  & k) \texttt{LRASPP}, $\mathcal{L}_{imp}$ & l) \texttt{LR-ASPP}, $\mathcal{L}_{entropy}$ \\ 

        \includegraphics[width=0.23\textwidth]{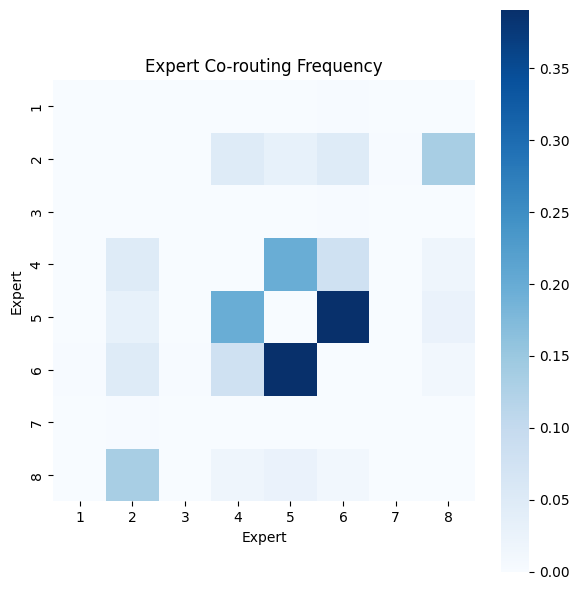}&
        \includegraphics[width=0.23\textwidth]{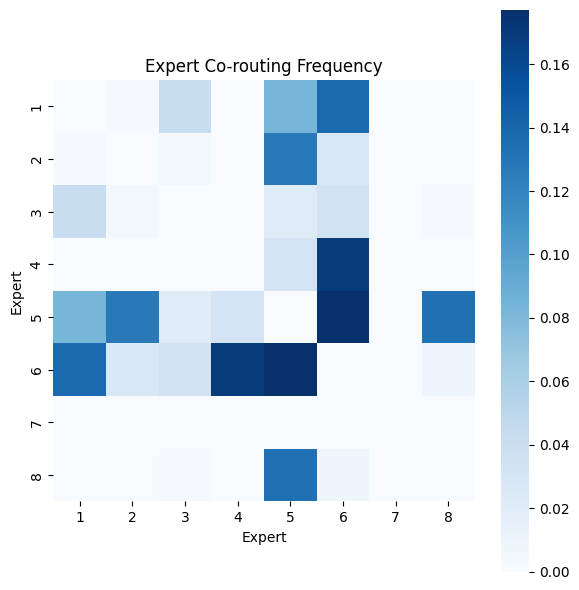}&
        \includegraphics[width=0.23\textwidth]{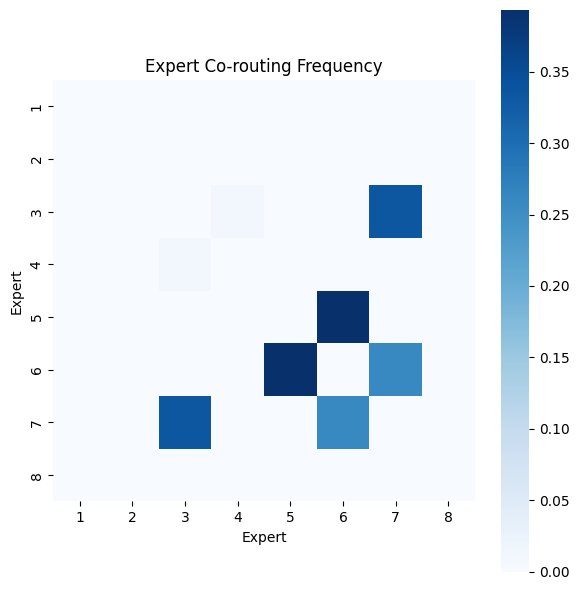}\\
        m) \texttt{DeepLabv3+}, $\mathcal{L}_{switch}$  & n) \texttt{DeepLabv3+}, $\mathcal{L}_{imp}$ & o) \texttt{DeepLabv3+}, $\mathcal{L}_{entropy}$ \\
    \end{tabular}
    }
    \caption{Expert co-routing frequency during inference. The settings follow Figure~\ref{fig:routing-heatmaps}.}
    \label{fig:co-routing}
\end{figure}

\textbf{Expert pair co-occurrence matrices} (see Figure~\ref{fig:co-routing}) show how often experts are activated together during top-$k$ routing. 
Across architectures, switch loss produces the most distributed co-activation with several active pairs and no dominant combination. 
Importance loss yields fewer but stable pairs, whereas entropy loss collapses to one or two recurring co-activations. 
Deeper models such as \texttt{ERFNet} and \texttt{DeepLabv3+} display the richest pair diversity under switch loss, while smaller networks like \texttt{ENet} show limited co-routing regardless of the loss. 
Overall, switch loss promotes broader expert collaboration, while importance and entropy lead to narrower, repetitive pair usage.

In summary, the extent of routing collapse depends strongly on both the balancing loss and model capacity. 
Backbone-based architectures show more stable and diverse expert usage, even with weaker balancing, suggesting that deeper feature representations support more reliable gating. 
Overall, higher model capacity and discriminative features mitigate collapse and promote broader expert usage.

\section{Conclusion}
In this work, we analyzed the behavior of sparse MoE layers in CNNs for semantic segmentation, focusing on routing stability and conditional computation. We adopted a patch-wise routing strategy that assigns local regions to a small subset of convolutional experts, enabling structured analysis of expert utilization and collapse. Across six architectures and two datasets, a single sparse MoE layer improved segmentation accuracy with minor parameter increases.

Extensive ablations of gating design, expert count, routing sparsity, and layer placement revealed consistent design trends. The most effective configuration used eight experts, top-2 routing, and a single MoE layer in the decoder, with balancing loss to mitigate collapse. Entropy loss often achieved the highest mIoU, while switch loss produced the most uniform expert usage and lowest routing collapse, while importance loss showed stronger expert dominance. Qualitative analyses further highlighted spatial and semantic specialization and its dependence on balancing strategy and model capacity.

Overall, this work provides empirical insight into the design and behavior of sparse MoE layers for CNN-based dense prediction and contributes to a better understanding of structured conditional computation in vision models.

\newpage
\clearpage
\section*{Acknowledgment}

This work was supported by funding from the Topic Engineering Secure Systems of the Helmholtz Association (HGF) and by KASTEL Security Research Labs (46.23.03).

{
    \small
    \bibliographystyle{ieeenat_fullname}
    \bibliography{references}
}


\end{document}